%% file: main.tex
\documentclass[12pt]{iopart}

\usepackage[utf8]{inputenc}
\usepackage[T1]{fontenc}
   \usepackage{microtype,newtxtext}

\usepackage{caption}
\usepackage{subcaption}% \usepackage[font=footnotesize,textfont=footnotesize]{subcaption}
\usepackage{multicol}
\usepackage{longtable,booktabs,array}
\expandafter\let\csname equation*\endcsname\relax
\expandafter\let\csname endequation*\endcsname\relax
\usepackage{amsmath}
\usepackage[switch]{lineno}
\usepackage{acronym}
\usepackage{graphicx}
\usepackage{xcolor}%[usenames,dvipsnames,svgnames,table]{xcolor}
\usepackage{xr} % for referring to the paper document
\usepackage{svg}
\usepackage[style=numeric-comp,backend=biber,maxcitenames=2,maxbibnames=1000,natbib=true]{biblatex}
\usepackage{hyperref}
\usepackage{soul}
\usepackage{lineno}
\graphicspath{{images/} {./} {Figures/experiments} {Figures/comparison} {Figures/mismatch}}
\DeclareGraphicsExtensions{.pdf,.png,.jpg,.mps,.svg}

\input{acronyms}

\begin{document}

\title[ARCANA]{A Realistic Simulation Framework for Analog/Digital Neuromorphic Architectures}
\author{Fernando M. Quintana$^{1,2,3\dagger}$\footnotetext{Corresponding author}, Maryada$^4$, Pedro L. Galindo$^3$, Elisa Donati$^4$, Giacomo Indiveri$^4$, Fernando Perez-Peña$^3$}

\address{$^1$ Bio-Inspired Circuits and Systems Lab, Zernike Institute for Advanced Materials, University of Groningen, Netherlands}
\address{$^2$ Groningen Cognitive Systems and Materials Center, University of Groningen, Netherlands}
\address{$^3$ School of Engineering, University of Cádiz, Puerto Real, Cádiz, Spain}
\address{$^4$ Institute of Neuroinformatics, University of Zurich and ETH Zurich, Zurich, Switzerland}
\ead{f.m.quintana.velazquez@rug.nl}
\vspace{10pt}
\begin{indented}
\item[]October 2024
\end{indented}

\begin{abstract}
  Developing dedicated mixed-signal neuromorphic computing systems optimized for real-time sensory-processing in extreme edge-computing applications requires time-consuming design, fabrication, and deployment of full-custom neuromorphic processors.
  To ensure that initial prototyping efforts exploring the properties of different network architectures and parameter settings lead to realistic results, it is important to use simulation frameworks that match as best as possible the properties of the final hardware.
  This is particularly challenging for neuromorphic hardware platforms made using mixed-signal analog/digital circuits, due to the variability and noise sensitivity of their components.
  In this paper, we address this challenge by developing a software spiking neural network simulator explicitly designed to account for the properties of mixed-signal neuromorphic circuits, including device mismatch variability.

 The simulator, called \textbf{ARCANA} (\textbf{A} \textbf{R}ealisti\textbf{c} Simulation Framework for \textbf{A}nalog/Digital \textbf{N}euromorphic \textbf{A}rchitectures), is designed to reproduce the dynamics of mixed-signal synapse and neuron electronic circuits with autogradient differentiation for parameter optimization and GPU acceleration.
  We demonstrate the effectiveness of this approach by matching software simulation results with measurements made from an existing neuromorphic processor.
  We show how the results obtained provide a reliable estimate of the behavior of the spiking neural network trained in software, once deployed in hardware.
  This framework enables the development and innovation of new learning rules and processing architectures in neuromorphic embedded systems.
\end{abstract}
\vspace{2pc}

\section{Introduction}
Mixed-signal neuromorphic circuits emulate the neural and synaptic dynamics observed in real neural systems, sharing similarities such as limited precision, heterogeneity, and high sensitivity to noise~\cite{Mead23,Chicca_etal14}, which are often overlooked in \ac{AI} workloads.
Typically, software simulations of \ac{SNN} use bit-precise activation functions identical for all neurons in a network with high-resolution parameters lacking biologically realistic details.
Conversely, tools used in computational neuroscience, such as “Neuron” and “NEST”, capture bio-realistic properties of neurons and synapses—such as ion channels and neuronal morphology~\cite{Hines_Carnevale97,Gewaltig_Diesmann07,Stimberg2019}.  In other words, these simulators attempt to replicate real hardware—in this case, \textit{the brain}—but do not offer modules for hyper-parameter optimization. Along similar lines, the neuromorphic community could benefit from a realistic simulator for electronic circuits that emulates the biophysics of neurons and synapses as well as the non-idealities of their analog substrates.

In this paper, we introduce a PyTorch-based~\cite{NEURIPS2019_9015} simulation platform, \textit{ARCANA}, for mixed-signal neuromorphic systems. ARCANA is optimized to simulate neurons and synapses using differential equations derived from their electronic circuits~\cite{Bartolozzi_Indiveri07a}. This seamless integration provides an easy-to-use infrastructure and libraries for parameter optimization, thereby facilitating the development of neural networks deployable on neuromorphic hardware.

This enables neuromorphic engineers, computational neuroscientists and \ac{AI} application developers to test and validate \ac{SNN} architectures in a fast prototyping environment that abstracts the underlying hardware intricacies~\cite{pehle2022brainscales,spilger2020hxtorch,yu2024integration,memtorch,aihwkit}.

A hardware-aware network can be effortlessly deployed on an inference chip, such as those from the DYNAP family~\cite{Moradi_etal18,Richter_etal24}. Moreover, the simulator can be readily adapted to incorporate hardware-specific constraints during model training, enabling a unified framework across various platforms. This approach offers two significant advantages: (1) it simplifies the transfer of a single model across multiple platforms, ensuring seamless cross-platform deployment, and (2) it facilitates benchmarking of different hardware using diverse datasets. Additionally, if a chip supports on-chip learning, deploying a pre-trained network enables continuous adaptation to new real-world data.

\section{Methods}
\label{sec:methods}
The mixed-signal circuit equations used in this framework are based on the \ac{DPI} circuit~\cite{Bartolozzi_Indiveri07a}, which is commonly employed to implement both analog synapse and neuron circuits~\cite{Chicca_etal14,Rubino_etal20,Lebanov_etal23,Mirshojaeian-Hosseini_etal22,Vuppunuthala_Pasupureddi23,Quan_etal23}.
The \ac{DPI} synapse equations faithfully reproduce realistic synaptic dynamics~\cite{Bartolozzi_Indiveri07a} and the corresponding neuron equations mimic the behavior of the  \ac{AdExpIF} neuron model~\cite{Brette_Gerstner05}.
Because the \ac{DPI} is a current-mode log-domain filter, both the synapse and the internal state variables of the neurons are expressed as \emph{currents} (e.g. membrane current), even though theoretical and computational models, typically represent the neuron’s state as the membrane \emph{potential}.
In the simulator, we adopt terminology and variable names that closely match those used in the electronic circuits.

The mixed-signal neuromorphic processor used to validate the simulations is the \ac{DYNAP-SE}~\cite{Moradi_etal18}. It is a multi-core~\ac{SNN} processor with four cores, each with 256~\ac{AdexLIF} neurons, providing a total of 1024 neurons~\citep{Moradi_etal18}.

\subsection{DPI synapse}
\label{sec:dpi_synapse}

The simulator models four different types of synapses: two excitatory synapses (AMPA and NMDA) and two inhibitory (GABA\textsubscript{A} and GABA\textsubscript{B}). They vary in the dynamics and how they interact with the somatic compartment. (1) \textbf{AMPA synapses} are composed of a DPI block and are connected to the input of the neuron, modeling biological AMPA receptor; (2) \textbf{NMDA synapses}, also an excitatory synapses and linked to neuron's input, however, unlike AMPA synapses, it incorporates a pair of differential blocks to introduce a voltage gating mechanism. This mechanism blocks the synaptic current until the neuron's membrane potential reaching a specific threshold (\verb|Inmda_thr| parameter in ARCANA); (3) \textbf{GABA\textsubscript{A}, an inhibitory synapses}, implements a current mirror to subtract presynaptic current from the input of the neuron, and (4) \textbf{GABA\textsubscript{B} synapses} has the current mirror directly connected to the somatic compartment, thus subtracting presynaptic current from membrane current, mimicking a shunting behavior.

The dynamic behavior of the DPI circuit can be approximated as a first-order differential in equation~\citep{Chicca_etal14b}:

\begin{equation}
\label{eq:dpisynapse}
\centering
    \tau \frac{d}{dt}I_{syn} + I_{syn} = \frac{I_{g}}{I_\tau}I_{w}
\end{equation}

where $\tau=CU_T/\kappa I_\tau$ is the synapse decay time constant, $C$ the synapse capacitance, $U_T$ the thermal voltage, $\kappa$ the transistor subthreshold slope factor, $I_\tau$ the leakage current, $I_{g}$ the \ac{DPI} filter gain, and $I_w$ the base weight current, and $I_{syn}$ the synaptic current. These parameters are tunable and shared among synapses of the same type within each core.

\subsection{The DPI neuron}
\label{sec:dpi_neuron}
We derived the equations for the neuron model by directly mapping its circuit components. The \ac{DPI} neuron circuit comprises four main blocks:

\begin{itemize}
    \item \textbf{Input \ac{DPI} Model Leak}: The module integrates the \ac{DPI} inputs from the synapses and the constant DC input current to charge the capacitor $C_{mem}$, which represents the neuron's leak conductance. It has a series of transistors that control the input current gain, amplitude of DC current, and leakage current that discharges the capacitor. The neuron's input current is composed of the constant DC current plus the synaptic currents aggregated from $AMPA$ and $NMDA$, as well as $GABA_A$. Additionally, the $GABA_B$ synapse is connected to the leakage current, directly discharging the capacitor.

    \item \textbf{\acf{AHP} block}: This block provides slow negative feedback that models spike frequency adaptation. When a postsynaptic spike occurs, it integrates the event into a recurrent negative \ac{AHP} current, which is subtracted from the input, effectively suppressing the activity of the neuron.

    \item \textbf{Positive Feedback and Spike Generation Block}: This block mimics activation and inactivation of sodium channels through a positive feedback circuit. When the neuron current starts to ramp up, the current driving the inverter circuits is fed back into the capacitor $C_{mem}$, further increasing the neuron's membrane current. Consequently, the $I_{mem}$ current grows exponentially until the inverter circuits finish switching, at which point a spike is generated and the reset block is activated.

    \item \textbf{Reset Block}: This block mimics the potassium channels. After a spike event, it resets the membrane current by creating a short circuit to ground, which discharges the neuron's membrane capacitance and directs the membrane current $I_{mem}$ to ground. This discharge period, controlled by a bias parameter, prevents $C_{mem}$ from recharging, resulting in an absolute reset and a refractory period. Once this period ends, the $I_{mem}$ current recharges the capacitor $C_{mem}$ and the neuron resumes integrating its input spikes.

\end{itemize}

The dynamics of these modules are governed by the following differential equations \citep{Chicca_etal14b}:

\begin{subequations}\label{eq:dpineuron}
\begin{align}
&\left(1+\frac{I_{g}}{I_{mem}}\right)\tau\frac{d I_{mem}}{dt}
    + I_{mem}\left(1+\frac{I_{ahp}}{I_\tau}\right)
    = I_\infty + f(I_{mem})\\[1ex]
&\tau_{ahp}\frac{d I_{ahp}}{dt} + I_{ahp}
    = I_{ahp_\infty}\,u(t)\\[1ex]
&I_\infty
    = \frac{I_{g}}{I_\tau}\Bigl(I_{in} - I_{ahp} - I_\tau\Bigr)
\end{align}
\end{subequations}

where $I_{mem}$ denotes the subthreshold membrane current. $I_{ahp}$ is the after-hyperpolarization current responsible of spike-frequency adaptation. $I_\infty$ is the maximum current a neuron would reach asymptotically. $I_\tau$ represents leakage current of the neuron, $\tau$ is the neuron time constant, and $I_{in}$ is the total input current from both synaptic sources and the constant DC input. All these parameters can be configured both on \ac{DYNAP-SE} chip as well as the simulator.

The term $f(I_{mem})$ in equation \ref{eq:dpineuron}a represents the positive feedback current as described in section \ref{sec:dpi_neuron}, and it is well modeled by an exponential function.~\citep{Indiveri_etal10}:

\begin{equation}
\label{eq:pfb}
\centering
    f(I_{mem}) = \frac{I_{fb}}{I_{\tau}}(I_{mem} - I_{g}) \qquad
    I_{fb} = \frac{I_0^{\frac{1}{\kappa+1}}I_{mem}^{\frac{\kappa}{\kappa+1}}}{1 + e^{-\alpha(I_{mem} - I_{th})}}
\end{equation}

where $\alpha$ and $I_{th}$ are hyper-parameters, $I_0$ is dark current and $\kappa$ is transistor slope factor.

\paragraph{Parameters:} The digital-to-analog (DAC) bias-generator circuits configure bias parameters by generating specific magnitudes of current to govern neuronal and synaptic properties, including time constant, refractory period, and synaptic weights~\cite{Delbruck_Van-Schaik05}.
Internally, these parameters are represented as ``coarse'' and a ``fine'' value, where $coarse \in [0, 7]$ and $fine \in [0,255]$. These parameters exhibit variability, also known as mismatch, due to fabrication non-idealities, which inherently limits user control (see Section \ref{sec:mismatch}). Moreover, the assigned values are non-observable and cannot be directly retrieved from the circuits, so we rely on oscilloscope traces for measurement. The bias parameters that can be set on simulation are listed in Table \ref{tab:parameters}. In the case of $I_{th}$ and $\alpha$ they are fitting parameters of the positive feedback and not a direct bias in DYNAP-SE.

\begin{table}[]
\centering
    \caption{\label{tab:parameters}ARCANA model parameter sand DYNAP-SE biases correspondence.}
    \begin{tabular}{|c|c|c|}
    \br
    ARCANA & DYNAP-SE bias & Description\\
    \mr
    Igain\_mem & \verb|IF_THR_N| & neuron gain\\
    Itau\_mem & \verb|IF_TAU1_N|  & Membrane time constant\\
    refractory\footnotemark[1] & \verb|IF_RFR_N|  & refractory period\\
    Idc & \verb|IF_DC_P|& DC input \\
    Itau\_ampa & \verb|NPDPIE_TAU_F_P| & AMPA synaptic time constant\\
    Igain\_ampa &  \verb|NPDPIE_THR_F_P| & AMPA synaptic gain \\
    Iw\_ampa & \verb|PS_WEIGHT_EXC_F_N| & AMPA synaptic base weight current \\
    Inmda\_thr & \verb|IF_NMDA_N| & NMDA voltage gating\\
    Itau\_nmda &  \verb|NPDPIE_TAU_S_P| & NMDA synaptic time constant \\
    Igain\_nmda & \verb|NPDPIE_THR_S_P| & NMDA synaptic gain \\
    Iw\_nmda & \verb|PS_WEIGHT_EXC_S_N| & NMDA synaptic base weight current \\
    Itau\_gabaa & \verb|NPDPII_TAU_F_P| & GABA\_A synaptic time constant \\
    Igain\_gabaa & \verb|NPDPII_THR_F_P| & GABA\_A synaptic gain \\
    Iw\_gabaa & \verb|PS_WEIGHT_INH_F_N| & GABA\_A synaptic base weight current\\
    Itau\_gabab & \verb|NPDPII_TAU_S_P| & GABA\_B synaptic time constant \\
    Igain\_gabab & \verb|NPDPII_THR_S_P| & GABA\_B synaptic gain\\
    Iw\_gabab & \verb|PS_WEIGHT_INH_S_N| & GABA\_B synaptic base weight current \\
    \br
    \end{tabular}\\
    \footnotesize{\footnotemark[1] The refractory period in ARCANA is in milliseconds compared to the bias current in DYNAP-SE.}
\end{table}

\subsection{Calibration}
\label{sec:calib}
We performed a one-time calibration (per chip) to align the simulation parameters (in pA) with those on the chip (in a.u.) using incremental adjustments. We primary calibrated the membrane leak time constant, $I_\tau$, because it is the only bias that directly influences the membrane current \textemdash a quantity measurable by an oscilloscope. All other parameters were kept fixed to ensure consistency during the calibration process.

\begin{figure}[t]
\centering
\caption{ARCANA and \ac{DYNAP-SE} comparison, where on each figure, the \textit{left pane} represents the voltage and the \textit{right pane} the soma current.}
\begin{subfigure}[b]{\textwidth}
\includegraphics[width=\linewidth]{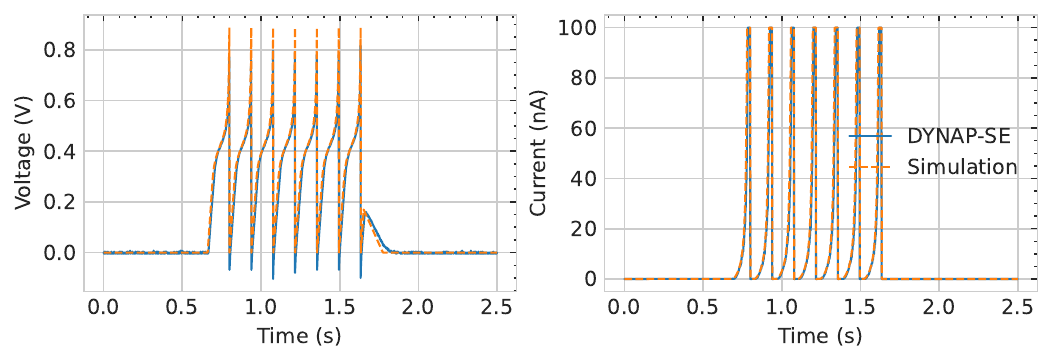}
\caption{Using a constant DC input, we compared ARCANA simulator with the \ac{DYNAP-SE} chip . For visualization purposes, the current was capped at $100\ nA$, although in practice it would rise exponentially until the firing threshold was reached.}
\label{fig:comparison:constant_DC}
\end{subfigure}

\begin{subfigure}[b]{\textwidth}
\includegraphics[width=\linewidth]{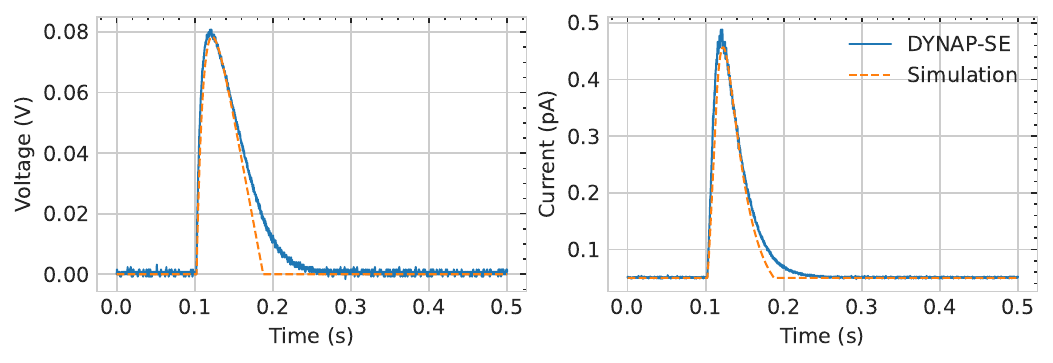}
\caption{Comparison between ARCANA and \ac{DYNAP-SE} chip using a AMPA input synapse. The traces exhibit a mean squared error ($MSE$) of $1.888 \times 10^{-7}$.}
\label{fig:comparison:AMPA_syn}
\end{subfigure}

\begin{subfigure}[b]{\textwidth}

\includegraphics[width=\linewidth]{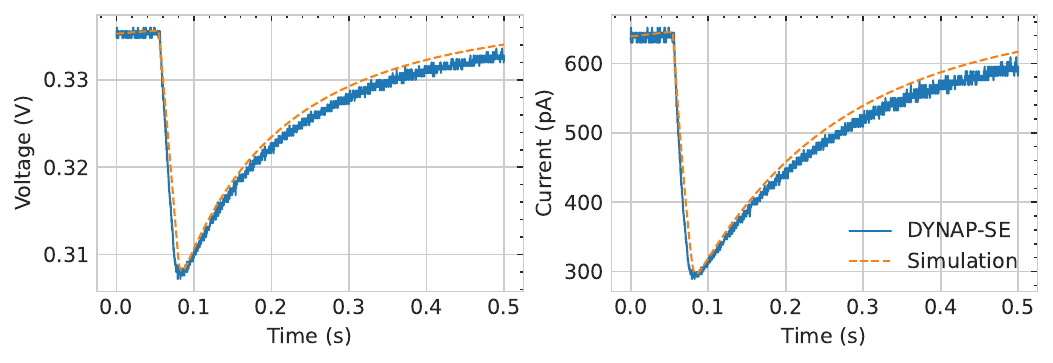}
\caption{Comparison between ARCANA and \ac{DYNAP-SE} chip using a GABA\textsubscript{a} input synapse. The traces exhibit a mean squared error ($MSE$) of $1.879 \times 10^{-6}$.}
\label{fig:comparison:gaba_a_syn}
\end{subfigure}

\label{fig:comparison}
\end{figure}

Similarly, we also evaluated the above-threshold neural response by increasing the DC input ($I_{DC}$) and neuron gain ($I_{th}$).  A slight adjustment of the leak current ($I_\tau =4.1pA$) was sufficient to match the simulation and silicon traces, reducing the neuron's decay time constant. This calibration resulted in the values: $I_{th} = 500pA$, $I_\tau = 4.1pA$, and $I_{DC} = 36.6pA$ for simulation. These current values are obtained by setting a [coarse,fine] pair that controls the on-chip DAC current-bias generator~\cite{Delbruck_Van-Schaik05}: $I_{th} = [6,88]$, $I_\tau = [6,22]$, and $I_\tau = [2,57]$. Figure  \ref{fig:comparison:constant_DC} shows a close match between the simulation traces (dashed orange line) and the silicon neuron traces (solid blue line), confirming that the simulation equations accurately capture the parameters and that their influence on silicon neuron behavior is faithfully modeled.

Following a similar approach, we calibrated the synapse parameters by recording the neuron's membrane current in response to a single spike. For the AMPA synapse, the membrane current trace (Figure \ref{fig:comparison:AMPA_syn}) was obtained with $I_\tau = 8pA$ ($[coarse = 2, fine =20]$), $I_{th} = 6.5pA$ ($[coarse = 2, fine =20]$) and $I_{w} = 50pA$ ($[coarse = 5, fine =99]$). For the GABA\textsubscript{a} inhibitory synapse (Figure \ref{fig:comparison:gaba_a_syn}), we set $I_\tau = 20pA$ ($[coarse = 2, fine =80]$), $I_{th} = 350pA$ ($[coarse = 4, fine =80]$), $I_{DC} = 6.5pA$ ($[coarse = 2, fine = 10]$) and $I_{w} = 400pA$($[coarse = 5, fine =200]$) . Note that for inhibitory synapse, the neuron's membrane current must be above its resting level to observe the inhibitory effect.

We observed a higher deviation in traces for AMPA ($MSE = 1.888 \times 10^{-7}$) and GABA\textsubscript{A} synapses ($MSE = 1.879 \times 10^{-6}$) compared to the traces with constant input current. This discrepancy may have arisen from the scale of the simulated voltage. When the input is provided via a synapse, the somatic voltage only reaches about $80\ mV$, whereas with constant DC current, it can rise to $800\ mV$ due to a positive feedback mechanism. Another possible explanation is that parasitic capacitance from the circuit layout and fabrication, which is not accounted for in the simulation, could contribute to the difference.

Here, we calibrated a single neuron such that all others remain within a 20\% margin of variability caused by mismatch. A more rigorous approach would involve a statistical analysis of the entire population, as described in ~\cite{Zendrikov_etal23} using chip-in-loop calibration, but this is beyond the scope of the present work.

This calibration process is integrated in subsequent experiments (see sections: \ref{sec:image-classification} and \ref{sec:ELTP}). These experiments serve as validation tests, where a network trained in simulation was deployed on the chip and produced outcomes similar to those observed in ARCANA.

\section{Hardware mismatch}
\label{sec:mismatch}
Inherent variability in the analog substrate can cause computational discrepancies relative to idealized simulations, imposing constraints that are beyond the user's control. Typically, the coefficient of variation for all parameters ranges from $~ 10-20\%$~\cite{Zendrikov_etal23}.

\begin{figure}
    \centering
    \includegraphics[width=0.8\textwidth]{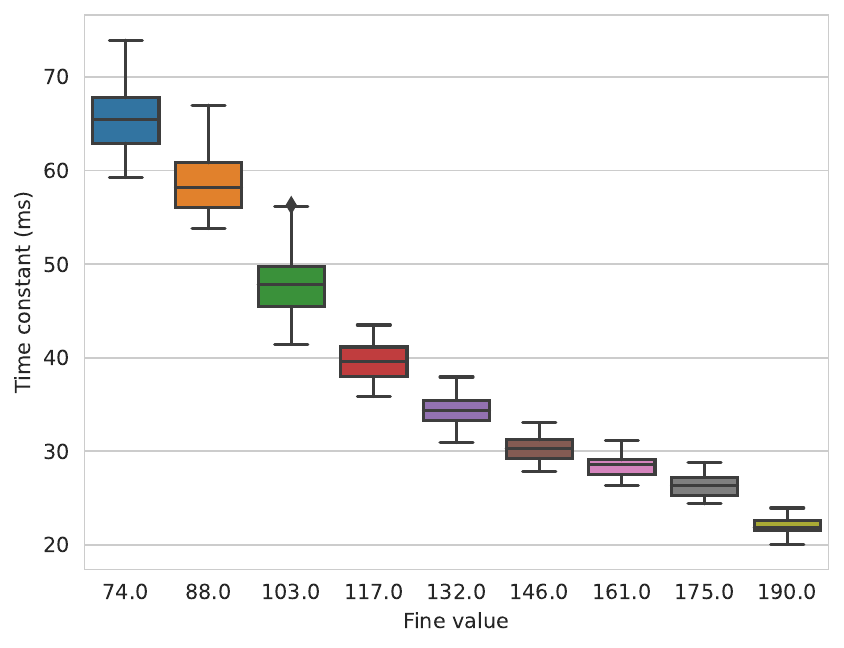}
\caption{Time constant distribution for $I_{\tau}$ ranging from 530 nA to 1676 nA.}
\label{fig:mismatch}
\end{figure}

This mismatch is evident while observing from membrane leak. Figure \ref{fig:mismatch} shows the distribution of time constant ($\tau_{mem}$) computed from membrane traces with bias settings applied via \verb|IF_TAU1_N|. The coarse bias is fixed at 5, while the fine bias is swept from 74 to 190. Notably, as the $\tau$ value is modified, its standard deviation decreases proportionally.

ARCANA incorporates mismatch in both neuron and synapse models, allowing users to assess their models' resilience to parameter variability. Users can select the desired degree of variability, starting with deterministic simulations for initial testing and gradually introducing mismatch as needed.

\section{Results}
\label{sec:result}
We present three experiments that assess the dynamics and accuracy of our PyTorch-based implementation of the DPI neuron and synapse model. The first experiment, described in Section \ref{sec:Spike-resonator}, features a frequency resonator used solely to validate the training process in simulation; it was not deployed on the chip. In all experiments, we employed the \emph{autograd} tool to automatically calculate gradients, thereby facilitating the tuning and optimization of complex models.

Once we validated the training process, we trained a small network to perform binary classification using \ac{BPTT}. The trained network was then deployed on \ac{DYNAP-SE} chip (see section: \ref{sec:image-classification}). In the last experiment (section: \ref{sec:ELTP}), we present a proof-of-concept for \ac{ETLP} \citep{Quintana2024}. In this experiment, a network was trained in simulation using \ac{ETLP} to distinguish between two spike patterns. After training, we deployed it on the \ac{DYNAP-SE} chip for inference.

These experiments serve as validation for the algorithm's compatibility with hardware-aware training. Deploying pre-learned weights and synaptic connections on mixed-signal hardware paves the way for innovative real-world applications in neuromorphic computing. Notably, before deployment, it is critical to perform a one-time manual calibration per chip, as detailed in Section \ref{sec:methods}. In future work, this calibration process can be automated using autograd and a chip-in-loop paradigm ~\cite{Maryada_etal23}.

\subsection{Spike frequency resonator}
\label{sec:Spike-resonator}

\begin{figure}
    \centering
    \includegraphics[width=\textwidth]{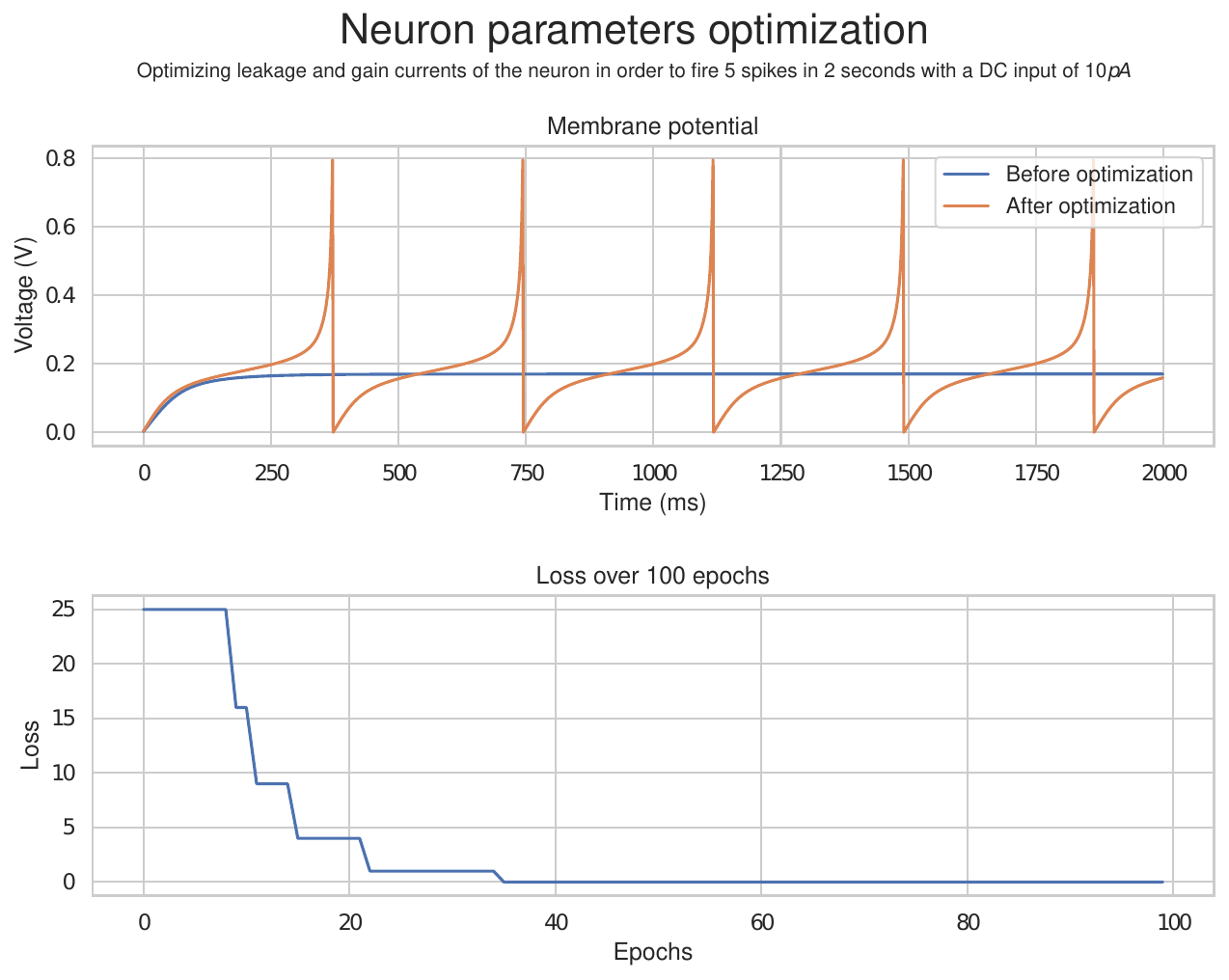}
    \caption{Neuron parameters optimization to obtain an output frequency of $2.5Hz$ from a constant input current of $10pA$.}
    \label{fig:voltageAdapt}
\end{figure}

In this experiment, we perform a basic task to showcase the use of the \emph{autograd} tool to optimize neuron parameters. The task involves constant current injection to a neuron with $I_{DC} =10 pA$. The objective is to modify neuron gain ($I_{th}$) and membrane time constant ($I_\tau$) to achieve the desired firing frequency of 2.5\,Hz. Figure \ref{fig:voltageAdapt} (top) shows the initial response of the neuron to DC injection (solid blue line). The absence of a spike response indicates the need to adjust the neuron's parameters. Instead of manually calibrating these parameters, we use autograd to automatically determine the optimal values so that the neuron fires at the user-specified frequency of 2.5\,HZ.

The loss function is the mean squared error between the number of output spikes and the target frequency. This loss is calculated with respect to the neuron parameters that we aim to optimize\textemdash neuron gain ($I_{th}$) and membrane time constant ($I_\tau$). Figure \ref{fig:voltageAdapt} (bottom) illustrates that the training was completed (loss $\simeq 0$) in fewer than 40 epochs. Figure \ref{fig:voltageAdapt} (top) also shows neuron's response post-optimization in orange.

\begin{table}
    \caption{\label{tab:spike_reso_param}Parameters used for spike frequency resonator experiment.}
    \begin{tabular*}{\textwidth}{@{}l*{15}{@{\extracolsep{0pt plus12pt}}l}}
        \br
        Parameter&Value \\
        \mr
        Initial neuron $I_{\tau}$ & $4e^{-12}$ \\
        Initial neuron $I_g$ & $20e^{-12}$ \\
        $I_{DC}$ & $10e^{-12}$ \\
        Positive feedback gain ($I_g$) & $20e^{-12}$ \\
        Positive feedback normalization ($\alpha$) & $2.0e^{9}$ \\
        Learning rate &  $5e^{-3}$ \\
        Optimizer & Adam \\
        \br
    \end{tabular*}
\end{table}

\subsection{Binary Image Classification}
\label{sec:image-classification}
\begin{figure}
    \centering
    \subcaptionbox{\label{fig:zerofreq}}{\includegraphics[width=0.49\textwidth]{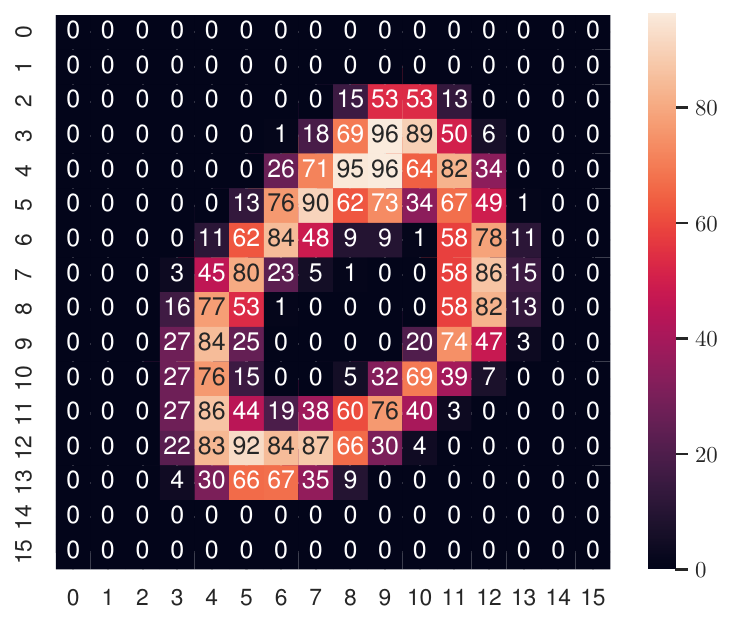}}
    \subcaptionbox{\label{fig:onefreq}}{\includegraphics[width=0.49\textwidth]{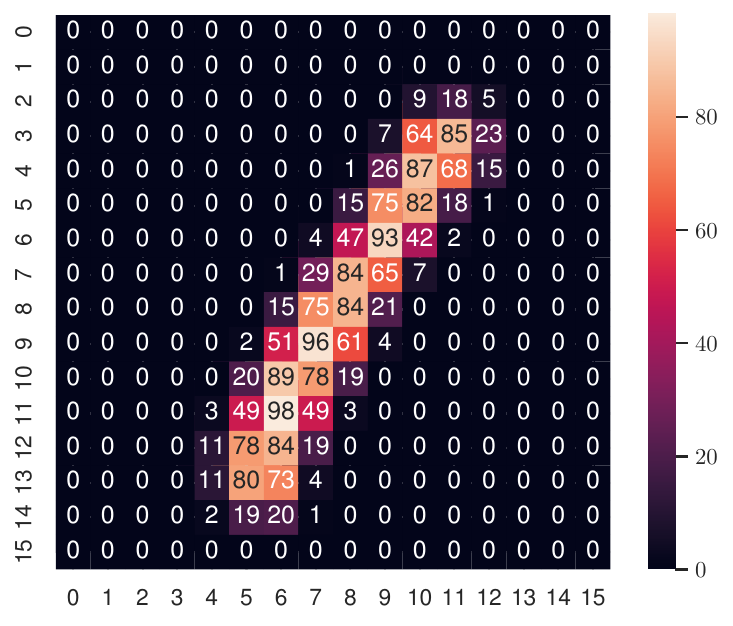}}
    \caption{Samples from class 0 (A) and 1 (B) with the corresponding frequency (Hz) value to convert it into a spike train.}
    \label{fig:freq}
\end{figure}

In this experiment, we trained the network to perform image classification for digit recognition. For simplicity, we only consider 0 and 1 of the MNIST dataset. We trained the connection matrix of the network and subsequently deployed the trained network on \ac{DYNAP-SE} for on-chip inference. The aim is to ensure that the designated neuron on the chip exhibits a higher firing rate than the baseline for its preferred digit (0 or 1). The network receives the pixel intensity encoded as firing rate with a Poisson distribution as shown in Figure \ref{fig:freq}.

\paragraph{\textbf{Syncing ARCANA and DYNAP-SE:}}
To ensure a faithful deployment of a trained network, it is essential to match the leakage and gain currents for both neurons and synapses on the chip and the ARCANA simulator, as described in section \ref{sec:calib}. On the \ac{DYNAP-SE} chip, the base weight for each synapse type can be tuned using bias currents, similar to other neuron and synaptic parameters (see section: \ref{sec:dpi_neuron}). This base weight is common to all neurons in a single core. Hence, to modify the weights for a specific neuron, we alter the number of synapses connecting two neurons (connection matrix). This necessitates the quantization of the learned connection matrix into integer values. We dealt with quantization constraint by employing \ac{QAT} procedure.

\paragraph{\textbf{Network training:}}

The classification task discussed here is linearly separable, therefore, we train a single-layer network with 256 input channels and 2 readout neurons, with one neuron representing digit 0 and the other representing digit 1. Each neuron can receive input with multiple excitatory (AMPA) and inhibitory (GABA\textsubscript{A}) synapses. We initialized AMPA and GABA\textsubscript{A} connection matrices with a uniform distribution. We pruned the synapses to meet the fan-in limitations (64 per neuron) of the chip.
We introduced mismatch in the synaptic weights bias ($I_w$) during training. However, all other neuron and synaptic parameters were manually calibrated (\ref{sec:mismatch}) and remain unchanged during training. Note that the framework also supports incorporating mismatch across all parameters during training, even though this feature is not utilized in this experiment.

Training with 32-bit floating-point weights provides a broader dynamic range but complicates on-chip deployment, as \ac{DYNAP-SE} uses quantized weights. With the help of quantization techniques such as \ac{QAT} \citep{nvidia2021quantization}, we reduced the precision of weights from float to integer, resulting in efficient computation while maintaining high accuracy during inference on the chip. A rounding operator was used to map the floating point tensor to a quantized representation:
$$x_q = round(x_f)$$

\begin{figure}[t]
\centering

\begin{subfigure}[b]{0.85\textwidth}
\includegraphics[width=\linewidth]{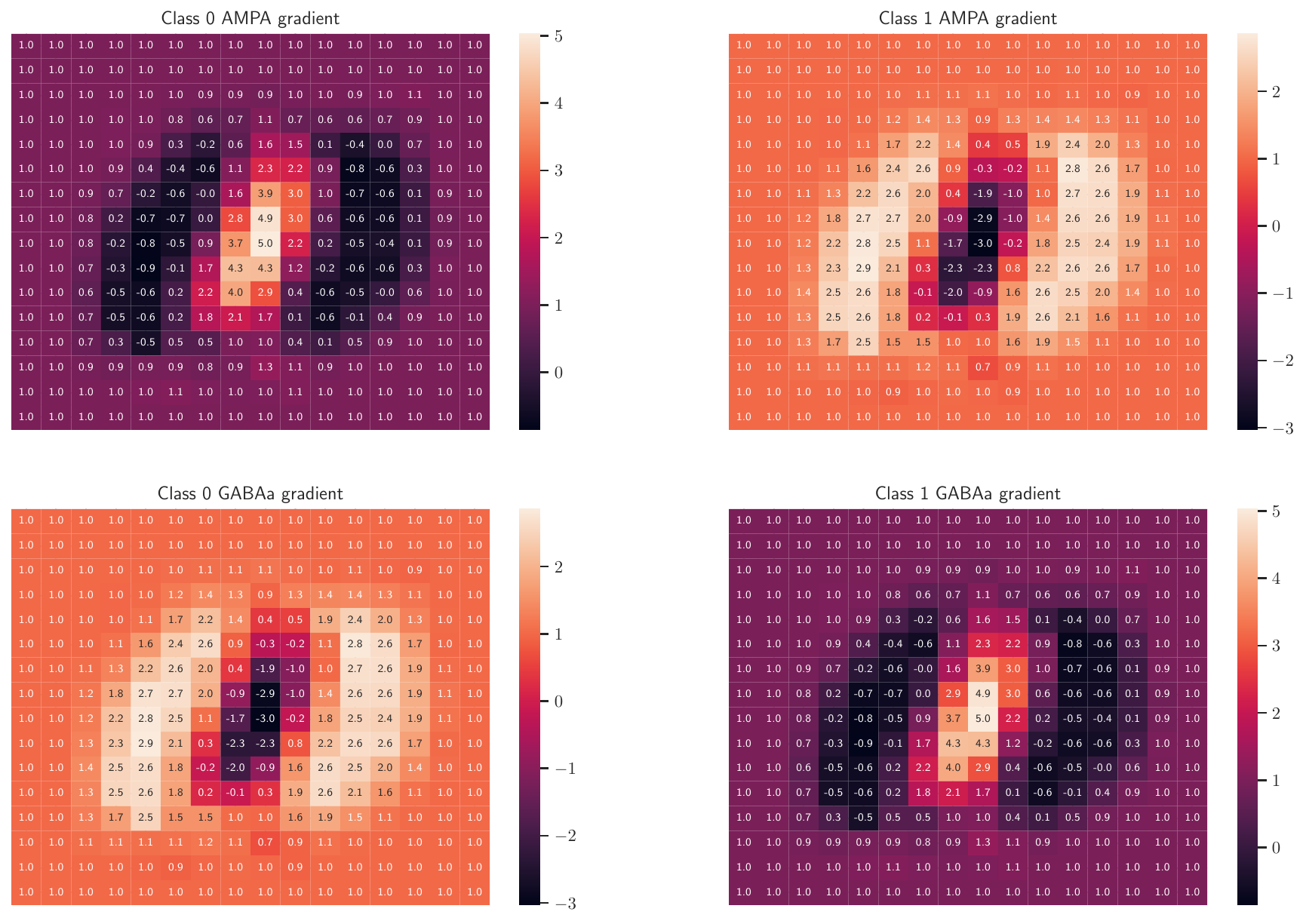}
\caption{AMPA and GABA\textsubscript{a} synapses gradients during the training process for classes 0 and 1. The value of the gradients are stored and calculated in full-precision.}
\label{fig:mnist:gradients}
\end{subfigure}

\begin{subfigure}[b]{0.85\textwidth}
\includegraphics[width=\linewidth]{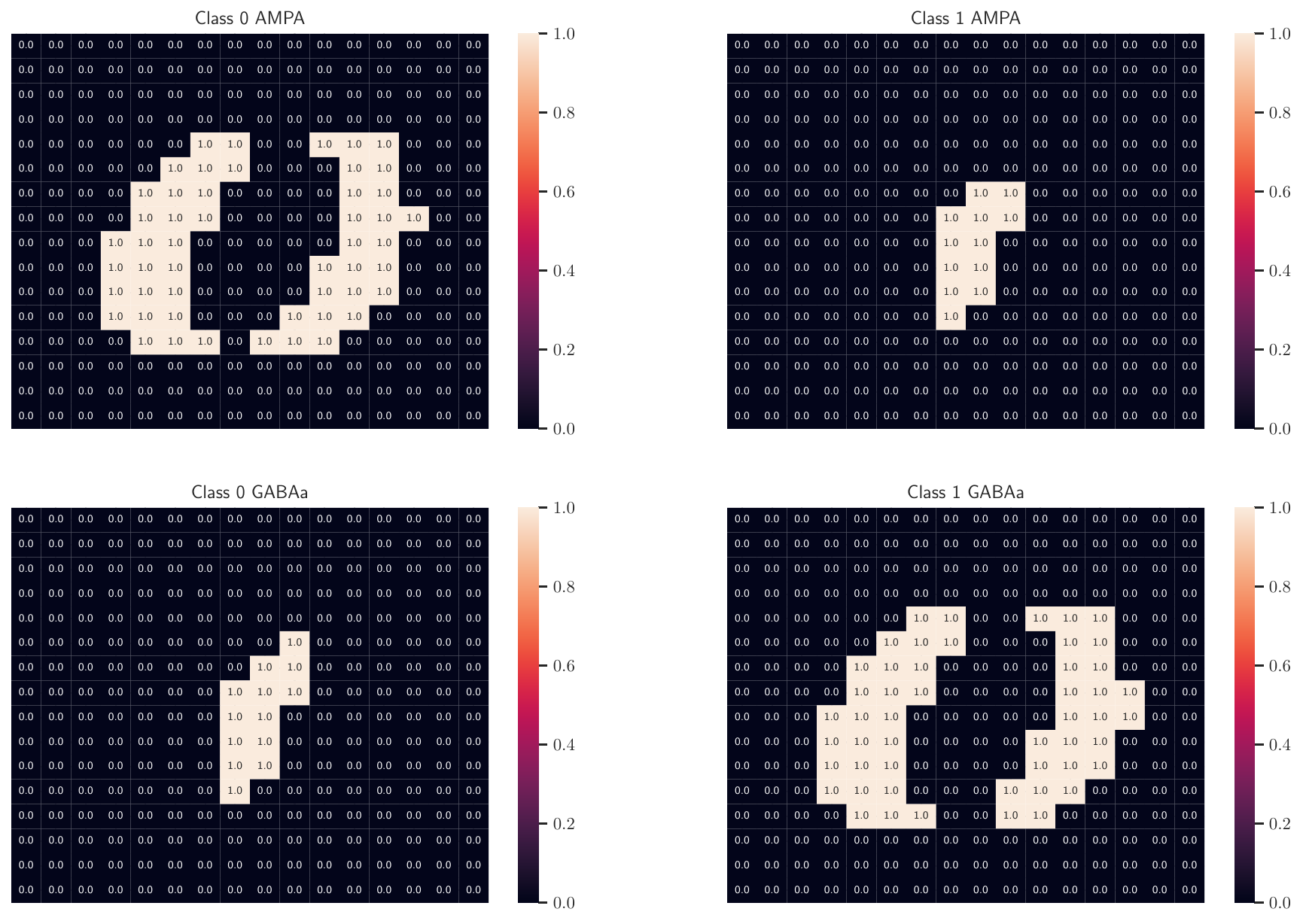}
\caption{AMPA and GABA\textsubscript{a} final connection matrix for class 0 and 1 obtained after training and quantizing them.}
\label{fig:weightMatrixRounded}
\end{subfigure}

\caption{AMPA and GABA\textsubscript{a} gradients and final quantized weight matrices.}
\label{fig:matrices}
\end{figure}

During training, \ac{QAT} introduces a ``mock'' low precision in the forward pass, while the backward pass remains full precision (Figure \ref{fig:mnist:gradients}). To deal with the quantization operation gradient, the \ac{STE} surrogate gradient is used. This approach allows the gradient to be transmitted unaltered through the fake-quantization operator.

As previously stated, the \ac{DYNAP-SE} chip also restricts the fan-in of a neuron, permitting only 64 input synapses per neuron. This restriction can differ for different chips, however it must be taken in account during training to avoid exceeding the synapse limit and risking a topology that is incompatible with the hardware. To address this challenge, we penalize the model for higher fan-in by adding L1 and/or L2 regularization terms. For instance, for $W$ as the connection matrix and $fan\_in$ as the desired number of input synapses (here $fan\_in = 64$ ), we used a L1 regularization, where L1 = $\lambda \sum|W - fan\_in|$. $\lambda$ is a hyperparameter that controls the strength of the regularization. If the sum of the rounded weights is not yet exactly equal to the desired value, a final adjustment can be made by adding a constant value $C$ to each weight, defined as $C=(fan\_in - \sum(round(W))) / N$. Here $N$ is the number of input neurons, and the sum is over all the input weights of each postsynaptic neuron.

\paragraph{\textbf{Loss:}} The loss function for this task is a softmax cross-entropy applied to the neuron inputs. Here, the network is optimized to increase the input current of the output neuron corresponding to the presented class while reducing the current of the other neuron. We defined the loss function on the total input current of each neuron (difference between the AMPA and GABA\textsubscript{A} current) in Equation \ref{eq:nmnistloss}.

\begin{subequations}
\begin{align}
    \label{eq:nmnistloss}
    L = -\sum^N_{n=1} \sum_{q=1}^Q y_{n,q}\,log\frac{e^{x_{n,q}}}{\sum_{i=1}^Q e^{x_{n,i}}}\\
    x_n = \sum_{t=0}^T I_{AMPA}(t)_n - I_{GABA_A}(t)_n \\
\end{align}
\end{subequations}

Where $y_{n,q}$ is the target class probability for sample $n$ and class $q$, $x_n$ is the difference between the accumulation over time of the AMPA and GABA\textsubscript{a} currents over time, $Q$ the number of classes, $N$ the number of samples, and $T$ the total simulation time.

\begin{figure}
    \centering
    \begin{subfigure}[b]{0.80\textwidth}
        \includegraphics[width=\linewidth]{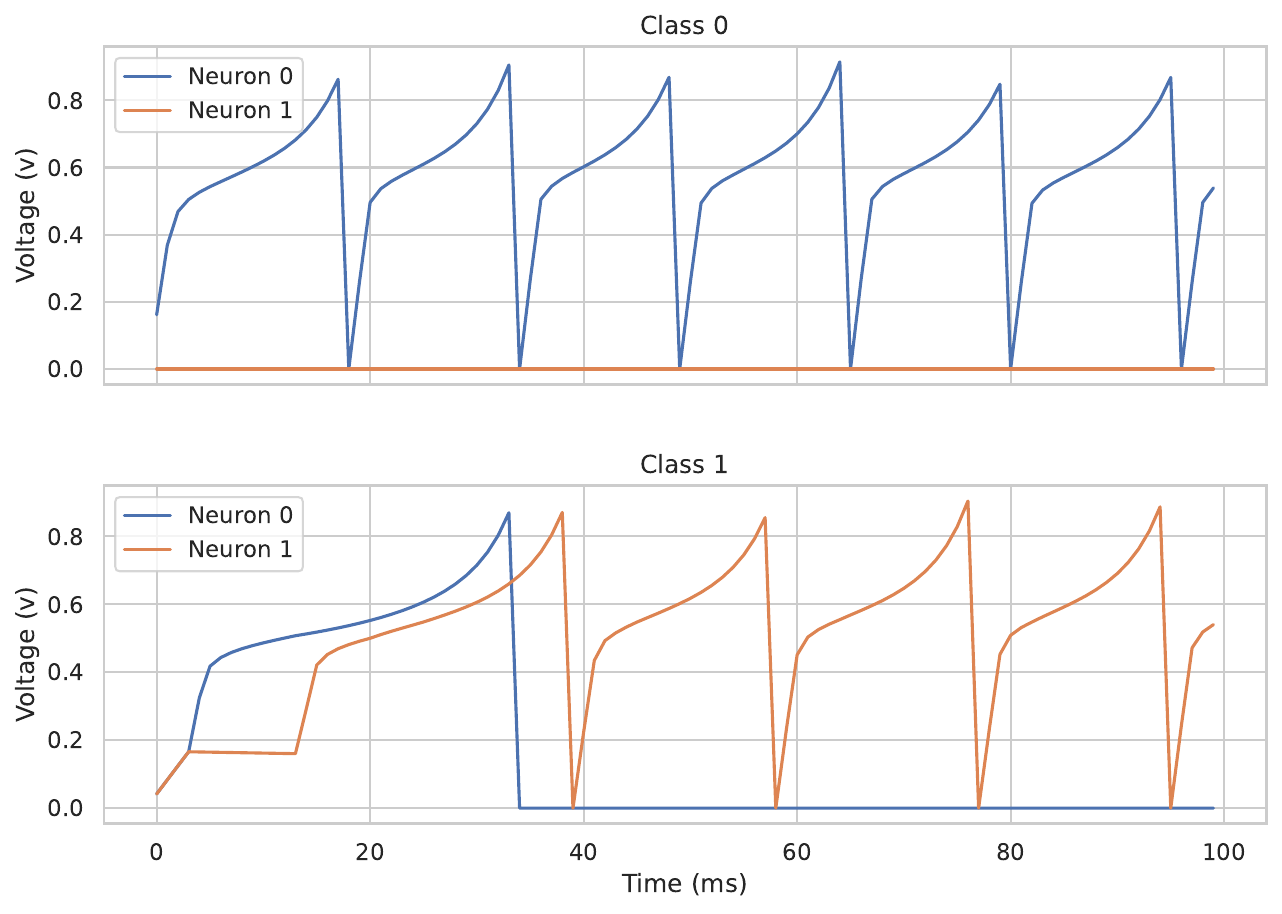}
        \caption{Output neurons voltages simulated in ARCANA when receiving a sample from class 0 and class 1.}
        \label{fig:mnist:voltages:simulation}
    \end{subfigure}

    \begin{subfigure}[b]{0.80\textwidth}
        \centering
        \includegraphics[width=\textwidth]{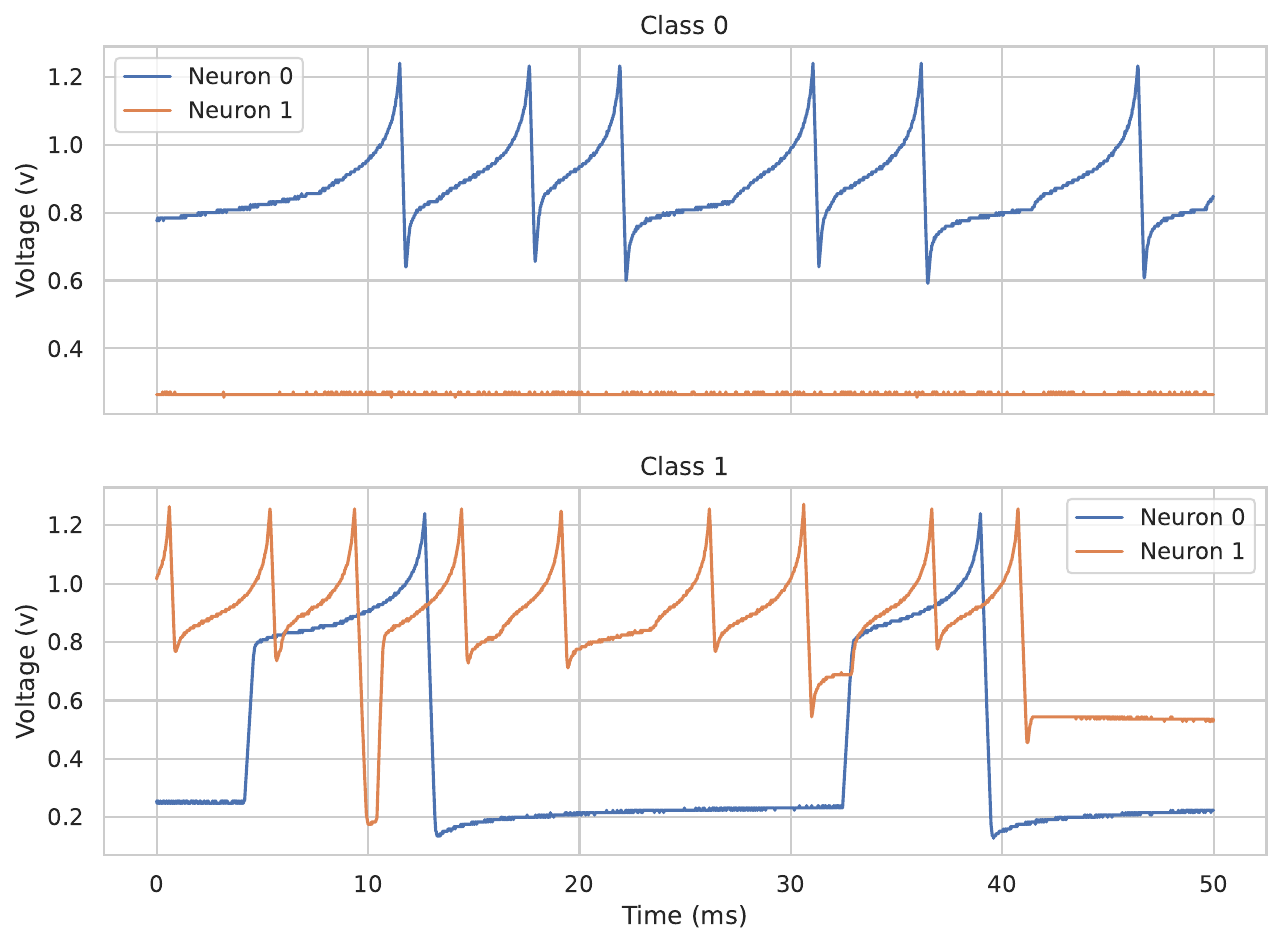}
        \caption{Output neurons voltages recorded on \ac{DYNAP-SE} when receiving a sample from class 0 and class 1. The figure shows the output of the neuron voltage recorded directly from the hardware, from a sample of each class.}
        \label{fig:mnist:voltages:dynapse}
    \end{subfigure}
    \caption{Output neuron voltages for classes 0 and 1 recorded in simulation (a) and on chip (b).}
\end{figure}

The final connection matrix obtained after training is then rounded, as shown in Figure \ref{fig:weightMatrixRounded}. Before deploying the network on the DYNAP-SE, we performed an inference mode evaluation in simulation (Figure \ref{fig:mnist:voltages:simulation}). As anticipated, the neurons spike behavior only when exposed to their respective selective digits.

The network was deployed on the hardware as a last step in this process. During on-chip inference, 2115 test samples of MNIST digits (0 or 1) were randomly selected and presented to the network for 50 ms each, interleaved with 50 ms rest periods with no stimulus. The firing response of the neurons was recorded for each presentation. Figure \ref{fig:mnist:voltages:dynapse} shows representative traces recorded on the chip during one inference session. Running entirely on hardware, the emulated network achieved an accuracy of \textit{99.11\%} in predicting the stimulus. This demonstrates the feasibility of training network connections in software and efficiently deploying it on a mixed-signal chip.

\begin{table}
    \caption{\label{tab:mnist_params}Parameters used for binary classification problem.}
    \begin{tabular*}{\textwidth}{@{}l*{15}{@{\extracolsep{0pt plus12pt}}l}}
        \br
        Parameter&Value \\
        \mr
        Neuron $I_{\tau}$ & $1.8e^{-12}$ \\
        Neuron $I_g$ & $45e^{-12}$ \\
        $I_{DC}$ & $240e^{-12}$ \\
        Positive feedback gain ($I_g$) & $225e^{-12}$ \\
        Positive feedback normalization ($\alpha$) & $1.0e^9$ \\
        GABA\textsubscript{a}/AMPA synapse $I_{\tau}$ & $4e^{-12}$ \\
        GABA\textsubscript{a}/AMPA synapse $I_g$ & $10e^{-12}$ \\
        GABA\textsubscript{a}/AMPA synapse base weight & $400e^{-12}$ \\
        Learning rate &  $1e^{-3}$ \\
        Optimizer & SGD \\
        Maximum number of input synapses & 40 \\
        \br
    \end{tabular*}
\end{table}

\subsection{Learning rules for DYNAP-SE}
\label{sec:ELTP}
In this experiment, we showed how ARCANA is a promising tool-chain for end-to-end network training and on-chip inference post-deployment. An important advantage of having a simulator for a mixed-signal processor such as \ac{DYNAP-SE} is the possibility of testing various learning rules such as \ac{ETLP} \cite{Quintana2024} with the neuron and synapse model implemented on the hardware. Thus, it serves as a proof-of-concept to test the viability of these learning algorithms for specific hardware before designing the circuit layout of the learning rule for on-chip implementation.

\begin{figure}
    \centering
    \includegraphics[width=\textwidth]{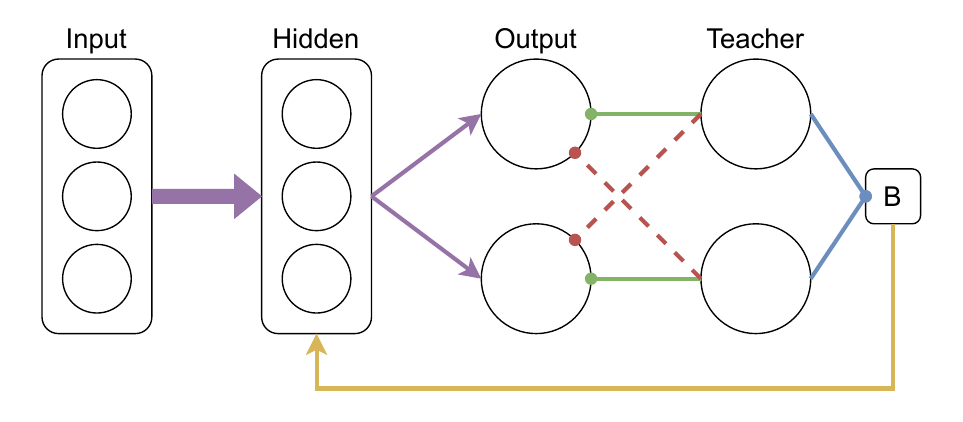}
    \caption{Architecture of ETLP learning rule. The teaching neurons are connected to the output neurons via excitatory (green) and inhibitory synapses (dashed red) to learn the weights between the hidden and output layers. The teaching neurons are also connected through a random initialized weight matrix B to the hidden layer neurons to apply the learning signal to the weight matrix between the input and the hidden layer.}
    \label{fig:etlp_schem}
\end{figure}

In this experiment, we trained a network using \ac{ETLP}~\cite{Quintana2024} and deployed the final weights on DYNAP-SE. \ac{ETLP} is a supervised three-factor local learning rule composed of 1) a presynaptic trace, 2) a postsynaptic value, and 3) an external teaching signal from teacher neurons that triggers the learning (Figure \ref{fig:etlp_schem}). The weight update follows equation \ref{sec:ETLP_eq2}.

\begin{subequations}
\begin{align}
    \tau_{\epsilon}\frac{d\epsilon}{dt} &= \epsilon + x\\
    \Delta W &= \sigma'(v(t))\epsilon(t)I(t)
    \label{sec:ETLP_eq2}
\end{align}
\end{subequations}

where $\Delta W$ is the weight change, $\epsilon$ is the pre-synaptic trace computed by accumulating the input spikes, $\sigma'(v) = (1 + \alpha|v|)^{-2}$ is a non-linear function depending of the soma voltage $v(t)$ \cite{zenke2018superspike}, $x$ the input pre-synaptic spike, $\tau_{\epsilon}$ the pre-synaptic trace time constant and $I(t)$ the input from the teaching neurons.

In this experiment, we trained a network (as illustrated in \ref{fig:etlp_schem}) to distinguish between two input frequencies \textemdash1 Hz and 10 Hz per channel. This system can be applied to anomaly detection. The network architecture consists of 50 input neurons, 50 hidden neurons, and 50 output neurons, along with 2 teacher neurons. A high-frequency teacher signal at 50 Hz facilitates learning of the preferred class, while a low-frequency teacher at 5 Hz simulates noise. Synaptic weight changes follow the \ac{ETLP} rule. For inference, the trained weights were deployed on the \ac{DYNAP-SE} chip.

\begin{figure}
    \centering
    \includegraphics[width=\textwidth]{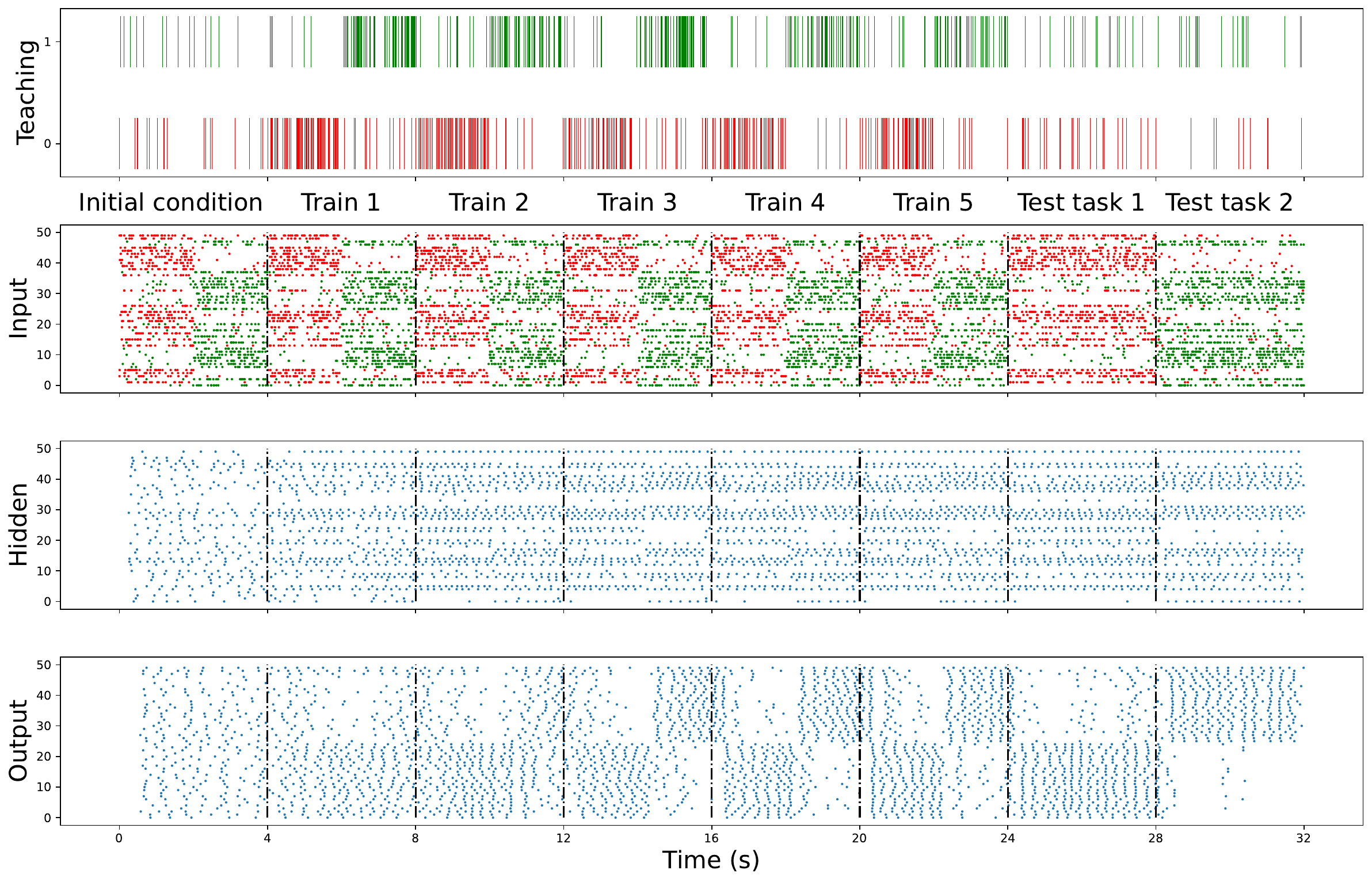}
    \caption{Raster plot of the simulation performed on ARCANA to train a network using \ac{ETLP}. The first plot represents the spikes of the teaching neurons that are connected to the hidden and output layers. The second plot corresponds to the input pattern, which has two neuron groups that fire with different frequencies. The network has to learn to associate each class with each neuron group. The third and fourth plots are the spikes produced by the hidden and output neurons during the training and testing processes.}
    \label{fig:etlp:software}
\end{figure}

\begin{table}
    \caption{\label{tab:ETLP_params}Parameters used for ETLP experiment. The presynaptic trace time constant is the same as the neuron time constant.}
    \begin{tabular*}{\textwidth}{@{}l*{15}{@{\extracolsep{0pt plus12pt}}l}}
        \br
        Parameter&Value \\
        \mr
        Neuron $I_{\tau}$ & $4.32e^{-12}$ \\
        Neuron $I_g$ & $20e^{-12}$ \\
        $I_{DC}$ & 0 \\
        Positive feedback gain ($I_g$) & $225e^{-12}$ \\
        Positive feedback normalization ($\alpha$) & $1.0e^9$ \\
        GABA\textsubscript{a}/AMPA synapse $I_{\tau}$ & $4e^{-12}$ \\
        GABA\textsubscript{a}/AMPA synapse $I_g$ & $10e^{-12}$ \\
        GABA\textsubscript{a}/AMPA synapse base weight & $200e^{-12}$ \\
        Maximum number of input synapses & 40 \\
        \br
    \end{tabular*}
\end{table}

Figure \ref{fig:etlp:software} shows spike activity of the whole network during the training and inference process in simulation. The output neurons fire independently for the initial 4 seconds of the pattern presentation. In the next 20 seconds, both tasks are presented alternatively in the network by increasing the firing rate of the input neurons from each class. This allows the hidden and output layers to learn the relation between the classes and the input pattern. During inference, the teacher neurons are set to a low-rate firing state (5 Hz), a subset of neurons in output layer exhibit higher firing response for their preferred class, validating correct association was learned during training.

\begin{figure}
    \centering
    \includegraphics[width=\textwidth]{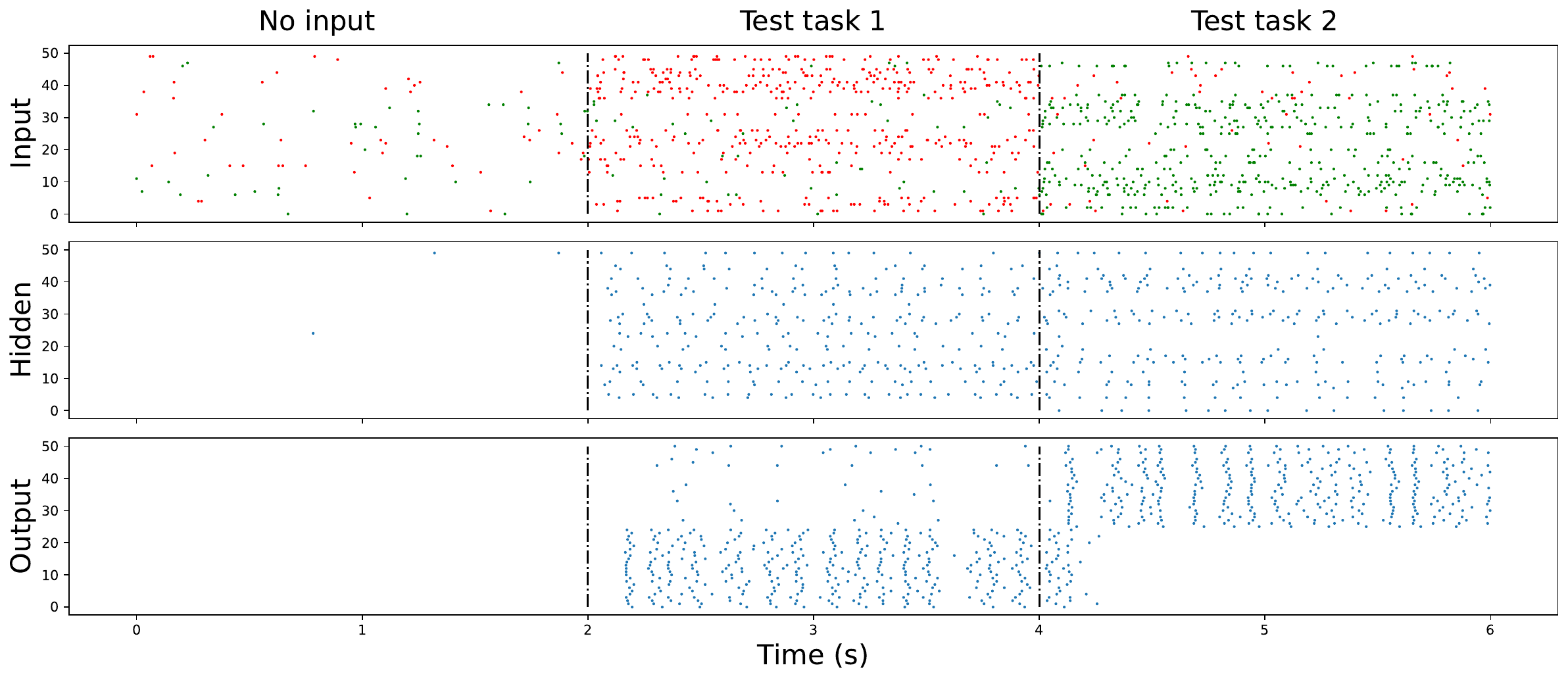}
    \caption{Raster plot of the network running on \ac{DYNAP-SE} chip with the weights obtained by training in simulation using \ac{ETLP}. The network was first training on simulation and then deployed to the chip.}
    \label{fig:etlp:dynapse}
\end{figure}

\paragraph{\textbf{On-chip inference:}} Once the network was trained and validated for inference in simulation, the trained weights were mapped into a network created on \ac{DYNAP-SE} chip without any preprocessing step. Figure \ref{fig:etlp:dynapse} shows the spike activity of the whole network on chip, exhibiting the desired behavior for each individual input. During on-chip inference, with no teacher provided, the bottom raster plot shows two distinct neuron groups, each exhibiting increased activity when their preferred pattern is presented.

\section{Discussion}
\label{sec:discussion}
In this paper, we presented the ARCANA simulator for mixed-signal hardware and validated its performance using the \ac{DYNAP-SE} chip. The advantage of ARCANA is its ability to use of the PyTorch ``autograd'' feature, which allows the optimization of the internal parameters of the network. While other similar simulators developed in parallel to this work have been proposed within the Rockpool framework~\cite{çakal2023training}, specifically tailored for DYNAP-SE2 chip~\cite{Richter_etal24}, ARCANA boasts a more versatile application and is generic to any processor incorporating \ac{DPI} based neuron models, independent of the specific simulation framework used. This adaptability of ARCANA is achieved by fine-tuning parameters such as the positive-feedback exponential function parameters and the chip constant values, including the transistor slope factor and neuron capacitor. Furthermore, ARCANA allows us to optimize not only the weights but other parameters as well. In addition, ARCANA includes the possibility to apply mismatch into the neuron and synapse parameters found in the hardware. This mismatch can be included in the training framework, making the system more robust to variability in the biases. It also provides a more realistic environment for simulation, where the non-idealities of the hardware are taken into account. Finally, we demonstrated the feasibility of implementing \ac{ETLP} in hardware and applied it to online training in mixed signal hardware.
This opens new doors to the development of new local learning rules in embedded systems that go beyond the ones presented so far~\cite{Khacef_etal23}.

\section*{Acknowledgements}
F.M.Q. was supported by FPU grant (FPU18/04321) from the Spanish Ministry of Universities.
This work is supported by the HORIZON EUROPE EIC Pathfinder Grant ELEGANCE (Grant No. 101161114), and has received funding from Swiss National Science Foundation (SNSF 200021E\_222393  ). This work has received funding from the Swiss State Secretariat for Education, Research and Innovation (SERI).
\section*{Competing interests}
The authors declare no competing interests.

\section*{Code availability}
\url{https://github.com/ferqui/ARCANA}
\printbibliography
\end{document}

%% file: acronyms.tex
\acrodef{AI}{Artificial Intelligence}
\acrodef{AdexLIF}[AdexLIF]{ Adaptive exponential Leaky Integrate-and-Fire}
\acrodef{AdExpIF}[AdExpI\&F]{ Adaptive Exponential Integrate \& Fire}
\acrodef{AHP}{After-HyperPolarization}
\acrodef{ALIF}[ALIF]{Adaptive Leaky-Integrate and Fire}
\acrodef{ANN}[ANN]{Artificial Neural Network}
\acrodef{ASIC}[ASIC]{Application-Specific Integrated Circuit}
\acrodef{BP}[BP]{Backpropagation}
\acrodef{BPTT}[BPTT]{Backpropagation Through Time}
\acrodef{DFA}{Direct Feedback Alignment}
\acrodef{DPI}[DPI]{Differential Pair Integrator}
\acrodef{DRTP}[DRTP]{Direct Random Target Projection}
\acrodef{DVS}{Dynamic Vision Sensor}
\acrodef{DYNAP-SE}{Dynamic Neuromorphic Asynchronous Processor}
\acrodef{ETLP}[ETLP]{Event-based Three-factor Local Plasticity}
\acrodef{FA}{Fast Adaptation}
\acrodef{FPGA}[FPGA]{Field-Programmable Gate Array}
\acrodef{IF}[IF]{Integrate and Fire}
\acrodef{LIF}[LIF]{Leaky Integrate-and-Fire}
\acrodef{LUT}[LUT]{Lookup table}
\acrodef{LR}{Learning Rate}
\acrodef{LTD}[LTD]{Long-Term Depression}
\acrodef{LTP}[LTP]{Long-Term Potentiation}
\acrodef{MAML}{Model-Agnostic Meta-Learning}
\acrodef{ML}[ML]{Machine Learning}
\acrodef{MSE}[MSE]{Mean Squared Error}
\acrodef{N-MNIST}[N-MNIST]{Neuromorphic-MNIST}
\acrodef{PTQ}{Post-Training Quantization }
\acrodef{QAT}{Quantization-Aware Training}
\acrodef{R-STDP}[R-STDP]{Reward-modulated Spike-Timing-Dependent Plasticity}
\acrodef{RNN}[RNN]{Recurrent Neural Network}
\acrodef{RSNN}[RSNN]{Recurrent Spiking Neural Network}
\acrodef{RTRL}[RTRL]{Real-Time Recurrent Learning}
\acrodef{STE}{Straight-Through Estimator}
\acrodef{SNN}[SNN]{Spiking Neural Network}
\acrodef{STDP}[STDP]{Spike-timing-dependent plasticity}

%% file: biblioncs.bib
@Article{Bartolozzi_Indiveri07a,
author		= {C. Bartolozzi and G. Indiveri},
title		= {Synaptic dynamics in analog {VLSI}},
journal		= {Neural Computation},
year		= {2007},
month		= oct,
volume		= {19},
number		= {10},
pages		= {2581--2603},
doi		= {10.1162/neco.2007.19.10.2581}
}

@Article{Brette_Gerstner05,
author		= {Brette, Romain and Gerstner, Wulfram},
title		= {Adaptive exponential integrate-and-fire model as an
		  effective description of neuronal activity},
journal		= {Journal of neurophysiology},
year		= {2005},
volume		= {94},
number		= {5},
pages		= {3637--3642},
publisher	= {American Physiological Society},
doi		= {10.1152/jn.00686.2005}
}

@Article{Chicca_etal14,
author		= {E. Chicca and F. Stefanini and C. Bartolozzi and G.
		  Indiveri},
title		= {Neuromorphic electronic circuits for building autonomous
		  cognitive systems},
journal		= {Proceedings of the {IEEE}},
year		= {2014},
month		= sep,
volume		= {102},
number		= {9},
pages		= {1367--1388},
issn		= {0018-9219},
doi		= {10.1109/JPROC.2014.2313954}
}

@Misc{DVS,
key		= {DVS},
title		= {Product of inilabs: Dynamic Vision Sensor},
howpublished	= {http://www.inilabs.com/products/dynamic-vision-sensors},
year		= {2009}
}

@Article{Delbruck_Van-Schaik05,
author		= {T. Delbruck and A. {Van Schaik}},
title		= {Bias Current Generators with Wide Dynamic Range},
journal		= {Analog Integrated Circuits and Signal Processing},
year		= {2005},
volume		= {43},
number		= {3},
pages		= {247--268}
}

@Article{Gewaltig_Diesmann07,
author		= {Marc-Oliver Gewaltig and Markus Diesmann},
title		= {{NEST} (NEural Simulation Tool)},
journal		= {Scholarpedia},
year		= {2007},
volume		= {2},
number		= {4},
pages		= {1430}
}

@Article{Hines_Carnevale97,
author		= {M.L. Hines and N.T. Carnevale},
title		= {The {NEURON} simulation environment},
journal		= {Neural Computation},
year		= {1997},
volume		= {9},
number		= {6},
pages		= {1179--1209},
publisher	= {MIT Press}
}

@InProceedings{Indiveri_etal10,
author		= {G. Indiveri and F. Stefanini and E. Chicca},
title		= {Spike-based learning with a generalized integrate and fire
		  silicon neuron},
booktitle	= {International Symposium on Circuits and Systems,
		  ({ISCAS})},
year		= {2010},
pages		= {1951--1954},
organization	= {IEEE},
address		= {Paris, France},
doi		= {10.1109/ISCAS.2010.5536980}
}

@Article{Khacef_etal23,
author		= {Khacef, Lyes and Klein, Philipp and Cartiglia, Matteo and
		  Rubino, Arianna and Indiveri, Giacomo and Chicca,
		  Elisabetta},
title		= {Spike-based local synaptic plasticity: a survey of
		  computational models and neuromorphic circuits},
journal		= {Neuromorphic Computing and Engineering},
year		= {2023},
month		= nov,
volume		= {3},
number		= {4},
pages		= {042001},
publisher	= {IOP Publishing},
issn		= {2634-4386},
doi		= {10.1088/2634-4386/ad05da},
url		= {http://dx.doi.org/10.1088/2634-4386/ad05da}
}

@Article{Lebanov_etal23,
author		= {Lebanov, Ana and Lopez, Mauricio Velazquez and De Roose,
		  Florian and Papadopoulos, Nikolas P and Indiveri, Giacomo
		  and Rubino, Arianna and Payvand, Melika and Smout, Steve
		  and Willegems, Myriam and Catthoor, Francky and others},
title		= {Flexible Unipolar {IGZO} Transistor-Based Integrate and
		  Fire Neurons for Spiking Neuromorphic Applications},
journal		= {Biomedical Circuits and Systems, {IEEE} Transactions on},
year		= {2023},
publisher	= {IEEE},
doi		= {https://doi.org/10.1109/TBCAS.2023.3321506}
}

@Article{Maryada_etal23,
author		= {Maryada, and Soldado-Magraner, Saray and Sorbaro, Martino
		  and Laje, Rodrigo and Buonomano, Dean V. and Indiveri,
		  Giacomo},
title		= {Stable recurrent dynamics in heterogeneous neuromorphic
		  computing systems using excitatory and inhibitory
		  plasticity},
year		= {2023},
month		= aug,
publisher	= {Cold Spring Harbor Laboratory},
doi		= {10.1101/2023.08.14.553298},
url		= {http://dx.doi.org/10.1101/2023.08.14.553298}
}

@Article{Mead23,
author		= {Mead, Carver},
title		= {Neuromorphic Engineering: In Memory of Misha Mahowald},
journal		= {Neural Computation},
year		= {2023},
volume		= {35},
pages		= {343--383},
doi		= {10.1162/neco_a_01553}
}

@Article{Mirshojaeian-Hosseini_etal22,
author		= {Mirshojaeian Hosseini, Mohammad Javad and Yang, Yi and
		  Prendergast, Aidan J and Donati, Elisa and Faezipour, Miad
		  and Indiveri, Giacomo and Nawrocki, Robert A},
title		= {An organic synaptic circuit: toward flexible and
		  biocompatible organic neuromorphic processing},
journal		= {Neuromorphic Computing and Engineering},
year		= {2022},
month		= sep,
volume		= {2},
number		= {3},
pages		= {034009},
publisher	= {IOP Publishing},
issn		= {2634-4386},
doi		= {10.1088/2634-4386/ac830c},
url		= {http://dx.doi.org/10.1088/2634-4386/ac830c}
}

@Article{Moradi_etal18,
author		= {Moradi, S. and Qiao, N. and Stefanini, F. and Indiveri,
		  G.},
title		= {A Scalable Multicore Architecture With Heterogeneous
		  Memory Structures for Dynamic Neuromorphic Asynchronous
		  Processors ({DYNAPs})},
journal		= {{IEEE} Transactions on Biomedical Circuits and Systems},
year		= {2018},
month		= feb,
volume		= {12},
number		= {1},
pages		= {106--122},
doi		= {10.1109/TBCAS.2017.2759700}
}

@Article{Quan_etal23,
author		= {Quan, Jiale and Liu, Zhen and Li, Bo and Luo, Jiajun},
title		= {Ultra-Low-Power Compact Neuron Circuit with Tunable
		  Spiking Frequency and High Robustness in 22 nm {FDSOI}},
journal		= {Electronics},
year		= {2023},
volume		= {12},
number		= {12},
publisher	= {MDPI AG},
issn		= {2079-9292},
doi		= {10.3390/electronics12122648},
url		= {http://dx.doi.org/10.3390/electronics12122648}
}

@Article{Richter_etal24,
author		= {Ole Richter and Chenxi Wu and Adrian M Whatley and German
		  K{\"o}stinger and Carsten Nielsen and Ning Qiao and Giacomo
		  Indiveri},
title		= {{DYNAP-SE2}: a scalable multi-core dynamic neuromorphic
		  asynchronous spiking neural network processor},
journal		= {Neuromorphic Computing and Engineering},
year		= {2024},
month		= jan,
volume		= {4},
number		= {1},
pages		= {014003},
publisher	= {IOP Publishing},
doi		= {10.1088/2634-4386/ad1cd7},
url		= {https://doi.org/10.1088/2634-4386/ad1cd7}
}

@Article{Rubino_etal20,
author		= {Rubino, Arianna and Livanelioglu, Can and Qiao, Ning and
		  Payvand, Melika and Indiveri, Giacomo},
title		= {Ultra-Low-Power {FDSOI} Neural Circuits for Extreme-Edge
		  Neuromorphic Intelligence},
journal		= {{IEEE} Transactions on Circuits and Systems I: Regular
		  Papers},
year		= {2020},
volume		= {68},
number		= {1},
pages		= {45--56},
publisher	= {IEEE},
doi		= {10.1109/TCSI.2020.3035575}
}

@Article{Vuppunuthala_Pasupureddi23,
author		= {Vuppunuthala, Srikanth and Pasupureddi, Vijay Shankar},
title		= {3.6-pJ/Spike, 30-Hz Silicon Neuron Circuit in 0.5-V, 65 nm
		  {CMOS} for Spiking Neural Networks},
journal		= {{IEEE} Transactions on Circuits and Systems {II}: Express
		  Briefs},
year		= {2023},
volume		= {},
number		= {},
pages		= {},
doi		= {10.1109/TCSII.2023.3324584}
}

@Article{Zendrikov_etal23,
author		= {Zendrikov, Dmitrii and Solinas, Sergio and Indiveri,
		  Giacomo},
title		= {Brain-inspired methods for achieving robust computation in
		  heterogeneous mixed-signal neuromorphic processing
		  systems},
journal		= {Neuromorphic Computing and Engineering},
year		= {2023},
month		= jul,
volume		= {3},
number		= {3},
pages		= {034002},
publisher	= {IOP Publishing},
issn		= {2634-4386},
doi		= {10.1088/2634-4386/ace64c},
url		= {http://dx.doi.org/10.1088/2634-4386/ace64c}
}


%% file: new.bib
@article{çakal2023training,
doi = {10.1088/2634-4386/ad2ec3},
url = {https://dx.doi.org/10.1088/2634-4386/ad2ec3},
year = {2024},
month = {3},
publisher = {IOP Publishing},
volume = {4},
number = {1},
pages = {014011},
author = {Ugurcan Cakal and  Maryada and Chenxi Wu and Ilkay Ulusoy and Dylan Richard Muir},
title = {Gradient-descent hardware-aware training and deployment for mixed-signal neuromorphic processors},
journal = {Neuromorphic Computing and Engineering },
}

@article{Chicca_etal14b,
  author  = {Elisabetta Chicca and Fabio Stefanini and Chiara Bartolozzi and Giacomo Indiveri},
  title   = {Neuromorphic electronic circuits for building autonomous cognitive systems},
  journal = {Proceedings of the {IEEE}},
  year    = {2014},
  volume  = {102},
  number  = {9},
  pages   = {1367--1388},
  doi     = {10.1109/JPROC.2014.2313954},
  issn    = {0018-9219}
}

@misc{nvidia2021quantization,
  title={Achieving FP32 Accuracy for INT8 Inference Using Quantization Aware Training with TensorRT},
  author={Neta Zmora and Hao Wu and Jay Rodge},
  publisher={Publisher},
  year={2021},
  month={7},
  howpublished={\url{https://developer.nvidia.com/blog/achieving-fp32-accuracy-for-int8-inference-using-quantization-aware-training-with-tensorrt/}}
}

@article{pehle2022brainscales,
  title={The BrainScaleS-2 accelerated neuromorphic system with hybrid plasticity},
  author={Pehle, Christian and Billaudelle, Sebastian and Cramer, Benjamin and Kaiser, Jakob and Schreiber, Korbinian and Stradmann, Yannik and Weis, Johannes and Leibfried, Aron and M{\"u}ller, Eric and Schemmel, Johannes},
  journal={Frontiers in Neuroscience},
  volume={16},
  pages={795876},
  year={2022},
  publisher={Frontiers Media SA}
}

@article{Quintana2024,
   author = {Fernando M Quintana and Fernando Perez-Peña and Pedro L Galindo and Emre O Neftci and Elisabetta Chicca and Lyes Khacef},
   doi = {10.1088/2634-4386/AD6733},
   issn = {2634-4386},
   issue = {3},
   journal = {Neuromorphic Computing and Engineering },
   month = {8},
   pages = {034006},
   publisher = {IOP Publishing},
   title = {ETLP: event-based three-factor local plasticity for online learning with neuromorphic hardware},
   volume = {4},
   url = {https://iopscience.iop.org/article/10.1088/2634-4386/ad6733 https://iopscience.iop.org/article/10.1088/2634-4386/ad6733/meta},
   year = {2024},
}

@inproceedings{spilger2020hxtorch,
  title={hxtorch: PyTorch for BrainScaleS-2: perceptrons on analog neuromorphic hardware},
  author={Spilger, Philipp and M{\"u}ller, Eric and Emmel, Arne and Leibfried, Aron and Mauch, Christian and Pehle, Christian and Weis, Johannes and Breitwieser, Oliver and Billaudelle, Sebastian and Schmitt, Sebastian and others},
  booktitle={IoT Streams for Data-Driven Predictive Maintenance and IoT, Edge, and Mobile for Embedded Machine Learning: Second International Workshop, IoT Streams 2020, and First International Workshop, ITEM 2020, Co-located with ECML/PKDD 2020, Ghent, Belgium, September 14-18, 2020, Revised Selected Papers 2},
  pages={189--200},
  year={2020},
  organization={Springer}
}

@article{Stimberg2019,
  author    = {Stimberg, Marcel and Brette, Romain and Goodman, Dan F.M.},
  doi       = {10.7554/eLife.47314},
  issn      = {2050084X},
  journal   = {eLife},
  month     = {8},
  pmid      = {31429824},
  publisher = {eLife Sciences Publications Ltd},
  title     = {{Brian 2, an intuitive and efficient neural simulator}},
  volume    = {8},
  year      = {2019}
}

@article{yu2024integration,
  title={Integration of physics-derived memristor models with machine learning frameworks},
  author={Yu, Zhenming and Menzel, Stephan and Strachan, John Paul and Neftci, Emre},
  journal={arXiv preprint arXiv:2403.06746},
  year={2024}
}

@INPROCEEDINGS{memtorch,

  author={Lammie, Corey and Azghadi, Mostafa Rahimi},

  booktitle={2020 IEEE International Symposium on Circuits and Systems (ISCAS)}, 

  title={MemTorch: A Simulation Framework for Deep Memristive Cross-Bar Architectures}, 

  year={2020},

  volume={},

  number={},

  pages={1-5},

  doi={10.1109/ISCAS45731.2020.9180810}}

@INPROCEEDINGS{aihwkit,

  author={Rasch, Malte J. and Moreda, Diego and Gokmen, Tayfun and Le Gallo, Manuel and Carta, Fabio and Goldberg, Cindy and El Maghraoui, Kaoutar and Sebastian, Abu and Narayanan, Vijay},

  booktitle={2021 IEEE 3rd International Conference on Artificial Intelligence Circuits and Systems (AICAS)}, 

  title={A Flexible and Fast PyTorch Toolkit for Simulating Training and Inference on Analog Crossbar Arrays}, 

  year={2021},

  volume={},

  number={},

  pages={1-4},

  doi={10.1109/AICAS51828.2021.9458494}}

@incollection{NEURIPS2019_9015,
title = {PyTorch: An Imperative Style, High-Performance Deep Learning Library},
author = {Paszke, Adam and Gross, Sam and Massa, Francisco and Lerer, Adam and Bradbury, James and Chanan, Gregory and Killeen, Trevor and Lin, Zeming and Gimelshein, Natalia and Antiga, Luca and Desmaison, Alban and Kopf, Andreas and Yang, Edward and DeVito, Zachary and Raison, Martin and Tejani, Alykhan and Chilamkurthy, Sasank and Steiner, Benoit and Fang, Lu and Bai, Junjie and Chintala, Soumith},
booktitle = {Advances in Neural Information Processing Systems 32},
pages = {8024--8035},
year = {2019},
publisher = {Curran Associates, Inc.},
url = {http://papers.neurips.cc/paper/9015-pytorch-an-imperative-style-high-performance-deep-learning-library.pdf}
}

@article{zenke2018superspike,
  title={Superspike: Supervised learning in multilayer spiking neural networks},
  author={Zenke, Friedemann and Ganguli, Surya},
  journal={Neural computation},
  volume={30},
  number={6},
  pages={1514--1541},
  year={2018},
  publisher={MIT Press One Rogers Street, Cambridge, MA 02142-1209, USA journals-info~…}
}
